\documentclass[11pt,a4paper]{article}
\usepackage[hyperref]{acl2021}
\usepackage{times}
\usepackage{latexsym}

\usepackage{CJKutf8}

\usepackage{booktabs} 
\usepackage{graphicx}
\usepackage{amsfonts}
\usepackage{amsmath}
 \usepackage{subfigure}
\usepackage[T1]{fontenc}
\usepackage{makecell}
\usepackage{multirow}
\usepackage{diagbox}
\usepackage{microtype}
\aclfinalcopy 
\def\aclpaperid{142} 

\title{Cross-Lingual Abstractive Summarization\\ with Limited Parallel Resources}

\author{Yu Bai, Yang Gao, Heyan Huang\thanks{~~Corresponding author.}  \\
    School of Computer Science and Technology,\\
    Beijing Institute of Technology, Beijing, China \\
    Southeast Academy of Information Technology, Fujian, China \\
    Beijing Engineering Research Center of High Volume Language Information \\ Processing and Cloud Computing Applications, Beijing, China\\

  \texttt{\{yubai,gyang,hhy63\}@bit.edu.cn} \\
}

\begin{document}
\maketitle
\begin{abstract}

Parallel cross-lingual summarization data is scarce, requiring models to better use the limited available cross-lingual resources. 
Existing methods to do so often adopt sequence-to-sequence networks with multi-task frameworks. 
Such approaches apply multiple decoders, each of which is utilized for a specific task.
However, these independent decoders share no parameters, hence fail to capture the relationships between the discrete phrases of summaries in different languages, breaking the connections in order to transfer the knowledge of the high-resource languages to low-resource languages.
To bridge these connections, we propose a novel Multi-Task framework for Cross-Lingual Abstractive Summarization (MCLAS) in a low-resource setting.
Employing one unified decoder to generate the sequential concatenation of monolingual and cross-lingual summaries, MCLAS makes the monolingual summarization task a prerequisite of the cross-lingual summarization (CLS) task. 
In this way, the shared decoder learns interactions involving alignments and summary patterns across languages, which encourages attaining knowledge transfer. Experiments on two CLS datasets demonstrate that our model significantly outperforms three baseline models in both low-resource and full-dataset scenarios. 
Moreover, in-depth analysis on the generated summaries and attention heads verifies that interactions are learned well using MCLAS, which benefits the CLS task under limited parallel resources.

\end{abstract}

\section{Introduction}
Cross-lingual summarization (CLS) helps people efficiently grasp  salient information from articles in a foreign language. 
Neural approaches to CLS require large scale datasets containing millions of cross-lingual document-summary pairs~\citep{zhu2019ncls,cao2020jointly, zhu2020attend}.
However, two challenges arise with these approaches:
1) most languages are low-resource, thereby lacking document-summary paired data; 2) large parallel datasets across different languages for neural-based CLS are rare and expensive, especially under the current trend of neural networks.
Therefore, a low-resource setting is more realistic, and challenging, one for cross-lingual summarization.
To our best knowledge, cross-lingual summarization under low-resource settings has not been well investigated and explored. Therefore, in this paper, we will develop a new model for cross-lingual abstractive summarization under limited supervision.

\begin{figure}[t]
\centering
\includegraphics[width=0.99\columnwidth]{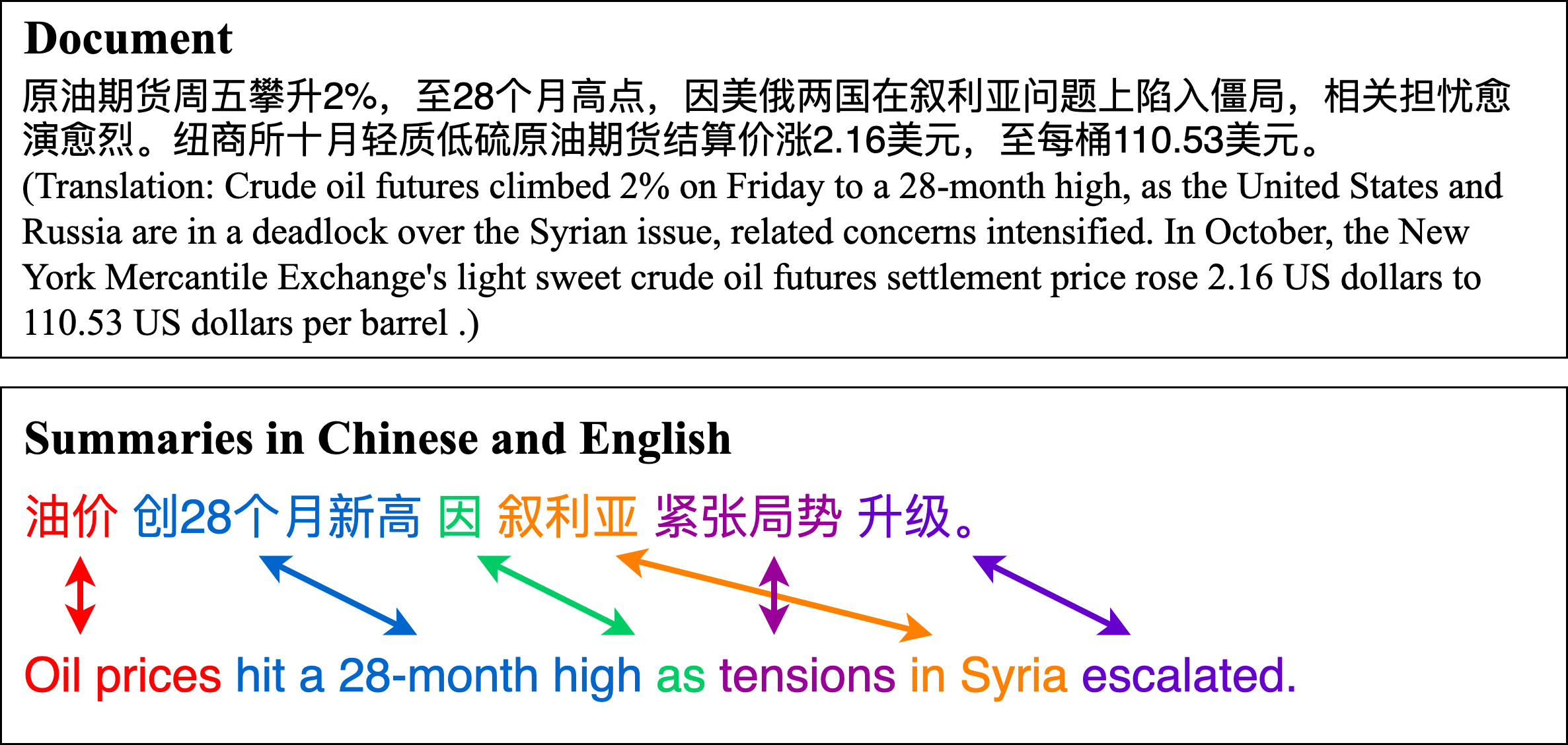} 
\caption{An example of the alignments across summaries in different languages. Each color represents phrases with one specific meaning.}
\label{fig:alignment}
\end{figure}

For low-resource settings, multi-task learning has been shown to be an effective method since it can borrow useful knowledge from other relevant tasks to use in the target task~\citep{yan2015multi,wang2020generalizing,motiian2017few}. 
Cross-lingual summarization can be viewed as the combination of two tasks, i.e., monolingual summarization (MS) and cross-lingual translation~\citep{zhu2019ncls}. 
A wealth of relationships exist across the target summaries of MS and CLS tasks, such as translation alignments and summarization patterns.
Illustrated in Figure~\ref{fig:alignment},
\begin{CJK*}{UTF8}{gbsn}
``叙利亚'' is mapped to ``Syria'',
and similar maping is done with the other aligned phrases.
\end{CJK*} 
Obviously, leveraging these relationships is crucial for the task of transferring summarization knowledge from high-resource languages to low-resource languages.
Unfortunately, existing multi-task frameworks simply utilize independent decoders to conduct MS and CLS task separately~\citep{zhu2019ncls, cao2020jointly},
which leads to failure in capturing these relationships. 

To solve this problem, we establish reliant connections between MS and CLS tasks, making the monolingual task a prerequisite for the cross-lingual task. 
Specifically, one decoder is shared by both MS and CLS tasks; this is done by setting the generation target as a sequential concatenation of a monolingual summary and the corresponding cross-lingual summary. 
Sequentially generating monolingual and cross-lingual summaries, the decoder also conducts the translation task between them, which enhances the interactions between different languages.
These \textbf{interactions} implicitly involve translation alignments,  similarity in semantic units, and summary patterns across different lingual summaries. To demonstrate these decoder interactions, we further visualize them by probing Transformer attention heads in the model. 
Based on this process, the new structure with these advanced interactions enhances low-resource scenarios which require the model to be capable of transferring summary  knowledge from high-resource languages to low-resource language. 
We name our model Multi-task Cross-Lingual Abstractive Summarization (MCLAS) under limited resources.

In terms of a training strategy under limited resources, 
we first pre-train MCLAS on large-scale monolingual document-summary parallel datasets to well-equip the decoder with general summary capability.   
Given a small amount of parallel cross-lingual summary samples, the model is then fine-tuned and is able to transfer the learned summary capability to the low-resource language, leveraging the interactions uncovered by the shared decoder.

Experiments on Zh2EnSum \citep{zhu2019ncls} and a newly developed En2DeSum dataset demonstrate that MCLAS offers significant improvements when compared with state-of-the-art cross-lingual summarization models in both low-resource scenarios and full-dataset scenario. 
At the same time, we also achieved competitive performances in the En2ZhSum dataset \citep{zhu2019ncls}. 
Human evaluation results show that MCLAS produces more fluent, concise and informative summaries than baselines models under limited parallel resources. 
In addition, we analyzed the length of generated summaries and the success of monolingual generation to verify advantages offered by identifying interactions between languages.
We further investigate the explainability of the proposed multi-task structure by probing the attention heads in the unified decoder, proving that MCLAS learns the alignments and interactions between two languages, and this facilitates translation and summarization in the decoder stage. Our analysis provides a clear explanation of why MCLAS is capable of supporting CLS under limited resources. Our implementation and data are available at \url{https://github.com/WoodenWhite/MCLAS}.

\section{Related Work}

\subsection{Cross-Lingual Summarization}
Recently, cross-lingual summarization has received attention in research due to the increasing demand to produce cross-lingual information.

Traditional CLS systems are based on a pipeline paradigm \cite{wan2010cross, wan2011using,zhang2016abstractive}. 
These pipeline systems first translate the document and then summarize it or vice versa.
\citet{shen2018zero} propose the use of pseudo summaries to train the cross-lingual abstractive summarization model. In contrast, \citet{duan2019zero} and \citet{ouyang-etal-2019-robust} generate pseudo sources to construct the cross-lingual summarization dataset. 

The first large-scale cross-lingual summarization datasets are acquired by use of a round-trip translation strategy \citep{zhu2019ncls}. 
Additionly, \citet{zhu2019ncls} propose a multi-task framework to improve their cross-lingual summarization system. 
Following~\citet{zhu2019ncls}, more methods have been proposed to improve the CLS task. \citet{zhu2020attend} use a pointer-generator network to exploit the translation patterns in cross-lingual summarization. \citet{cao2020jointly} utilize two encoders and two decoders to jointly learn to align and summarize. 
In contrast to previous methods, MCLAS generates the concatenation of monolingual and cross-lingual summaries, thereby modeling relationships between them.
\subsection{Low-Resource Natural Language Generation}
Natural language generation (NLG) for low-resource languages or domains has attracted lots of attention. \citet{gu2018meta} leverage meta-learning to improve low-resource neural machine translation. Meanwhile, many pretrained NLG models have been proposed and adapted to low-resource scenarios \citep{song2019mass, chi2020cross, radford2019language, zhang2019pegasus}. However, these models require large-scale pretraining. 
Our work does not require any large pretrained generation models or translation models, enabling a vital decreases in training cost.

\section{Background}
\label{sec:background}
\subsection{Neural Cross-lingual Summarization}
Given a source document $D^A = \{x_1^A, x_2^A, \dots, x_m^A\}$ in language $A$, a monolingual summarization system converts the source into a summary $S^A = \{y_1^A, y_2^A, \dots, y_n^A\}$,
where $m$ and $n$ are the lengths of $D^A$ and $S^A$, respectively.
A cross-lingual summarization system produces a summary $S^B = \{y_1^B, y_2^B, \dots, y_{n'}^B\} $ consisting of tokens $y^B$ in target language $B$, where $n'$ is the length of $S^B$. Note that the mentioned $x^A$, $y^A$, and $y^B$ are all tokens.

\citet{zhu2019ncls} propose using the Transformer~\citep{vaswani2017attention} to conduct cross-lingual summarization tasks. The Transformer is composed of stacked encoder and decoder layers. The encoder layer is comprised of a self-attention layer and a feed-forward layer. 
The decoder layer shares the same architecture as the encoder except for an extra encoder-decoder attention layer, which performs multi-head attention over the output of stacked encoder layers. The whole Transformer model $\theta$ is trained to maximize the conditional probability of the target sequence $S^B$ as follows:
\begin{equation}
\label{equ:nclsloss}
\small
    	L_{\rm NCLS} = \sum_{t=1}^{N}{\rm log} P(y^B_t|y^B_{<t},D^A)
\end{equation}

\subsection{Improving NCLS with Multi-Task Frameworks}
Considering the relationship between CLS and MS, in which they share the same goal to summarize important information in a document, \citet{zhu2019ncls} proposed employing a one-to-many multi-task framework to enhance the basic Transformer model. In this framework, one encoder is employed to encode the source document $D^A$. 
Two separate decoders simultaneously generate a monolingual summary $S^A$ and a cross-lingual summary $S^B$, leading to a loss as follows:
\begin{equation}
\label{equ:multitaskloss}
\small
\begin{array}{rl}
	L_{\rm NCLS+MS} = &  \sum_{t=1}^{n}{\rm log}P(y^{A}_t|y^{A}_{<t},D^A) \\
	& + \sum_{t=1}^{n'}{\rm log}P(y^{B}_t|y^{B}_{<t},D^A) \\
\end{array}
\end{equation}

This multi-task framework shares encoder representation to enhance cross-lingual summarization. However, independent decoders in this model are incapable of establishing alignments and connections between cross-lingual summaries.

\begin{figure}[t]
\centering
\includegraphics[width=0.99\columnwidth]{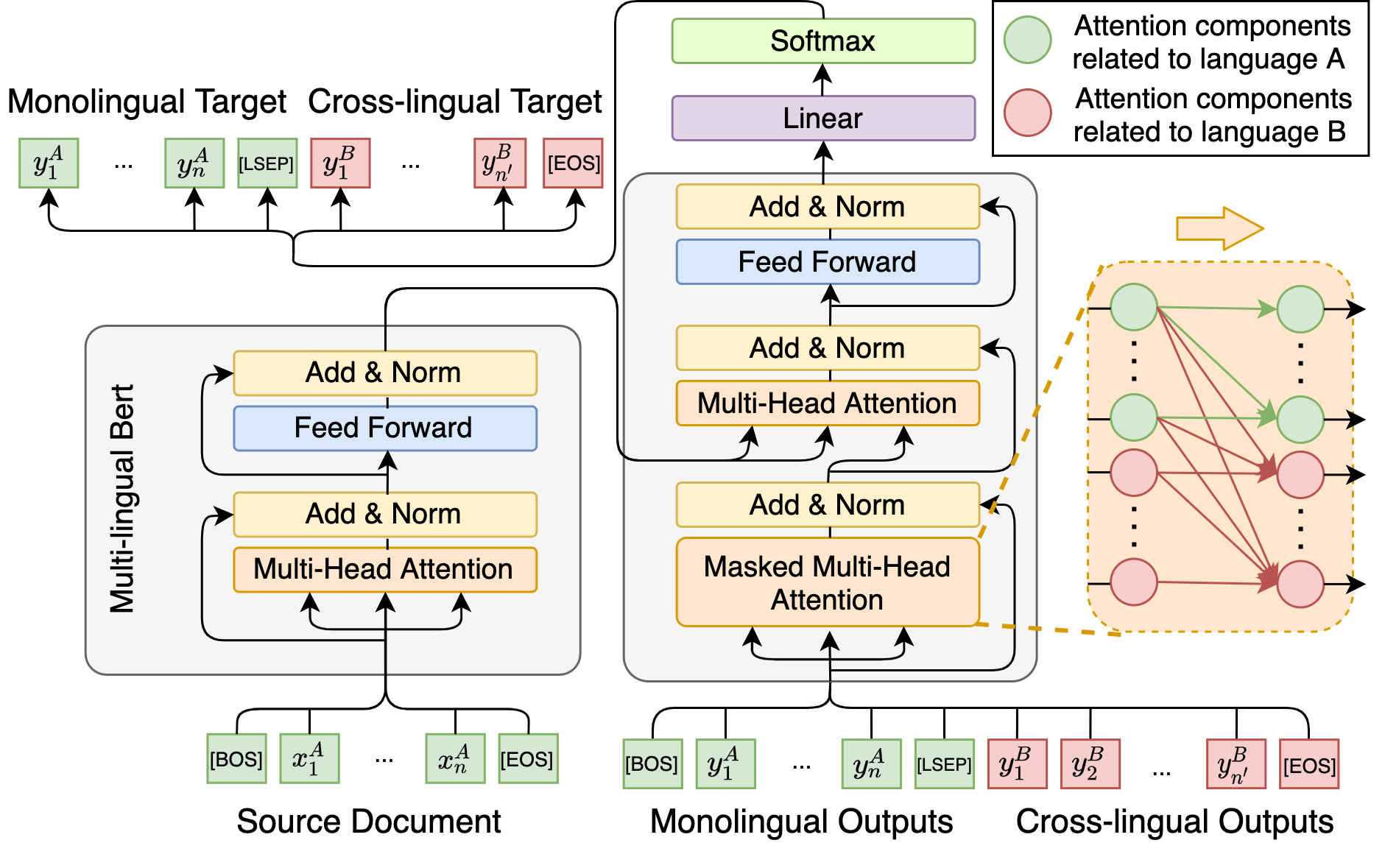} 
\caption{An overview of our proposed MCLAS. A unified decoder produces both monolingual (green) and cross-lingual (red) summaries. The green and red lines represent the monolingual and cross-lingual summaries' attention, respectively.}
\label{overall}
\end{figure}

\section{MCLAS with Limited Parallel Resources}
To strengthen the connections mentioned, we propose making the monolingual task a  prerequisite for the cross-lingual task through modeling interactions. 
According to previous work \citep{wan2010cross, yao2015phrase,zhang2016abstractive}, interactions between cross-lingual summaries (important phrase alignments, sentence lengths, and summary patterns, etc) are crucial for the final summary's quality. 
We leverage these interactions to further transfer the rich-resource language knowledge.
Detailed descriptions of this step are presented in following sections.

\subsection{Multi-Task Learning in MCLAS}
To model interactions between languages, we need to share the decoder's parameters.
Inspired by \citet{NEURIPS2019_c20bb2d9}, we propose sharing the whole decoder to carry out both the translation and the summarization tasks. 
Specifically, we substitute the generation target $S^A$ with the sequential concatenation of $S^A$ and $S^B$:
\begin{equation}
\small
\begin{array}{rl}
S^{AB} = & \{[\text{\textsc{bos}}], y_1^A, y_2^A, \dots, y_n^A, \\
 & [\text{\textsc{lsep}}], y_1^B, y_2^B, \dots, y_{n'}^B, [\text{\textsc{eos}}] \}\\
\end{array}
\label{Sab}
\end{equation}
where [\textsc{bos}] and [\textsc{eos}] are the beginning and end token of the output summaries, respectively. And [\textsc{lsep}] is the special token used as the separator of $S^A$ and $S^B$.

With the new generation target, the decoder learns to first generate $S^A$, and then generate $S^B$ conditioned on $S^A$ and $D^A$. The whole generation process is illustrated in Figure~\ref{overall}.

Formally, we maximize the joint probability for monolingual and cross-lingual summarization:
\begin{equation}
\label{equ:MCLASloss}
\small
\begin{array}{rl}
	L_{{\rm MCLAS}} = &  \sum_{t=1}^{n}{\rm log}P(y^{A}_t|y^{A}_{<t},D^A) \\
	& + \sum_{t=1}^{n'}{\rm log}P(y^{B}_t|y^{B}_{<t},S^A,D^A) \\
\end{array}
\end{equation}

The loss function can be divided into two terms. When generating $S^A$, the decoder conducts the MS task based on $D^A$, corresponding to the first term in Equation~(\ref{equ:MCLASloss}). 
When generating $S^B$, the decoder already knows the information of corresponding monolingual summaries. In this way, it performs the translation task (for $S^A$) and the CLS task (for $D^A$), achieved by optimizing the second term in Equation ~(\ref{equ:MCLASloss}). 
With the modification of the target, our model can easily capture interactions between cross-lingual summaries. The trained model shows effectiveness in aligning the summaries. Not only the output tokens, but also the attention distributions are aligned. The model we designed leverages this phenomenon to enable monolingual knowledge to be transferred under low-resource scenarios. Detailed investigation is presented in Section \ref{sec:probing}.

We adopt Transformers as our base model. 
In addition, we use multilingual BERT \cite{devlin2019bert} to initialize the encoder, improving its ability to produce multilingual representations.
Additionally, having tried many different position embedding and language segmentation embedding methods, we find that [\textsc{lsep}] is enough for the model to distinguish whether it is generating $S^B$. Hence keeping the original position embedding~\citep{vaswani2017attention} and employing no segmentation embedding are best for performance and efficiency.

\subsection{Learning Schemes for MCLAS under Limited Resources}
Since our proposed framework enforces interactions between cross multilingual summaries, it has further benefits to the low-resource scenario, as only a few training summary samples are available in a cross-language. 
Yet, simply training from scratch can not make the best of our proposed model in low-resource scenarios. 
Hence we use a pre-training and fine-tuning paradigm to transfer the rich-resource language knowledge. 
 
First, we train the model in a monolingual summarization dataset. In this step, the model learns how to produce a monolingual summary for a given document. Then, we jointly learn MS and CLS with few training samples, optimizing Equation~(\ref{equ:MCLASloss}). We adopt similar initialization to existing CLS methods, which is introduced in Section \ref{sec:baselines}.  

\section{Experiments}
\subsection{Datasets}
we conduct experiments on the En2ZhSum, Zh2EnSum CLS datasets\footnote{\url{www.nlpr.ia.ac.cn/cip/dataset.htm}}~\citep{zhu2019ncls} and a newly constructed En2DeSum dataset. 
En2ZhSum is an English-to-Chinese dataset containing 364,687 training samples, 3,000 validation, and 3,000 testing samples.   
The dataset is converted from the union set of CNN/DM \citep{hermann2015teaching} and MSMO \citep{zhu2018msmo} using a round-trip translation strategy. 
Converted from the LCSTS dataset, Zh2EnSum contains 1,693,713 Chinese-to-English training samples, 3,000 validation, and 3,000 testing samples.
To better verify the CLS ability of MCLAS,  we construct a new English-to-German dataset (En2DeSum), using the same methods proposed by \citet{zhu2019ncls}.
We use WMT'19 English-German winner\footnote{\url{https://github.com/pytorch/fairseq/tree/master/examples/translation}} as our translation model to process the English Gigaword dataset.\footnote{LDC2011T07} We set the threshold $T_1 = 0.6$ and $T_2=0.2$.
The final En2DeSum contains 429,393 training samples, 4,305 validation samples, and 4,099 testing samples.

\begin{table}[t]
\small
  \centering
  \resizebox{0.99\linewidth}{!}{
    \begin{tabular}{lrrr}
    \toprule
    Scenarios & Zh2EnSum & En2DeSum & En2ZhSum \\
    \midrule     
    Minimum& 5,000 (0.3\%) & 2,619 (0.6\%) & 1,500 (0.4\%) \\
    Medium  & 25,000 (1.5\%) & 12,925 (3.0\%) & 7,500 (2.0\%)\\
    Maximum & 50,000 (3.0\%) & 25,832 (6.0\%) & 15,000 (4.0\%) \\      
    Full-dataset &1,693,713  & 429,393 &  364,687 \\      
    \bottomrule
    \end{tabular}
    }
\caption{Sample sizes of different low-resource scenarios. Three low-resource scenarios with various sample sizes are created for each dataset. Minimum, Medium, and Maximum represent sample sizes in the minimum low-resource scenario, medium low-resource scenario, and maximum low-resource scenario, respectively.}
    \label{tab:few_size}
\end{table}

All the training samples contain a source document, a monolingual summary, and a cross-lingual summary. For the full-dataset scenario, we train the model with the whole dataset. 
For low-resource scenarios, we randomly select 3 different amounts (minimum, medium, and maximum) of training samples for all datasets to evaluate our model's performance under low-resource scenarios. Detailed numbers are presented in Table~\ref{tab:few_size}.

\begin{table*}[t]
\small
  \centering
  \resizebox{0.99\linewidth}{!}{
    \begin{tabular}{llrrrcrrrcrrrc}
    \toprule
    \multicolumn{2}{c}{\multirow{2}*{Models}} & \multicolumn{4}{c}{\textbf{Zh2EnSum}} & \multicolumn{4}{c}{\textbf{En2DeSum}} &
    \multicolumn{4}{c}{\textbf{En2ZhSum}} 
    \\
    \cmidrule(r{4pt}){3-6} \cmidrule(r{4pt}){7-10} \cmidrule(l){11-14}
    \multicolumn{2}{c}{~} & R-1 & R-2 & R-L & BERTScore& R-1 & R-2 & R-L & BERTScore & R-1 & R-2 & R-L & BERTScore\\
    \midrule
    \multirow{3}*{\makecell[l]{Minimum\\ Low-resource \\ Scenario}}  & NCLS & 20.93 & 5.88 & 17.58 & \textbf{0.5041} & 17.59 & 5.01&16.58& 0.7202 &\textbf{34.14} & 12.45 & \textbf{21.20} & 0.7096\\
    ~ & NCLS+MS  & 20.50  & 5.45 & 17.25 & 0.5025& 17.52  & 5.27 & 16.57 & 0.7198 & 33.96 & 12.38 & 21.07 & \textbf{0.7102} \\
    ~ & MCLAS & \textbf{21.03}  & \textbf{6.03 }& \textbf{18.16} & 0.5023 & \textbf{19.19}  & \textbf{5.91} & \textbf{18.43} & \textbf{0.7282}&32.03 &\textbf{ 13.17} & 21.17 & 0.6529\\
    \midrule
    \multirow{3}*{\makecell[l]{Medium \\ Low-resource \\ Scenario}}  & NCLS & 26.42 & 8.90 & 22.05 & 0.5373 & 23.55 & 8.09 & 22.13 & 0.7400 & 35.98 & 15.88 & 23.79 & \textbf{0.7298}\\
    
    ~ & NCLS+MS  & 26.86 & 9.06 & 22.47 & 0.5377& 23.60 & 8.35 & 22.14 & 0.7431 & 38.95 & 18.09 &\textbf{ 25.39} &0.7172 \\
    ~ & MCLAS & \textbf{27.84 }& \textbf{10.41} & \textbf{24.12} & \textbf{0.5464}& \textbf{27.22} & \textbf{10.09} & \textbf{26.00} & \textbf{0.7575} & 37.28 &\textbf{ 18.10} &25.26 & 0.6839 \\
    \midrule    
    \multirow{3}*{\makecell[l]{Maximum \\ Low-Resource \\ Scenario}}  & NCLS & 29.05 & 10.88 & 24.32 & 0.5492  & 25.84 & 9.78 & 24.25 & 0.7483 &\textbf{40.18}&19.86&26.52&0.7435  \\
    
    ~ & NCLS+MS  & 28.63 & 10.63 & 24.00 & 0.5485& 25.59 & 9.58 & 23.96 & 0.7484 &39.86 &\textbf{ 19.87} &\textbf{ 26.64} & \textbf{0.7445} \\
    ~ & MCLAS & \textbf{30.73} & \textbf{12.26} & \textbf{26.51} & \textbf{0.5633} & \textbf{30.31} & \textbf{12.32} & \textbf{28.88} & \textbf{0.7682} & 38.35 & 19.75 & 26.41 & 0.6921 \\
    \midrule
    \multirow{4}*{\makecell[l]{Full \\ dataset \\ Scenario}}  & TLTran & 33.64 & 15.58 & 29.74 & - & 28.57 & 13.31 &26.34 & - & 30.20 & 12.20 & 27.02 & -\\
    ~ & NCLS & 35.60 & 16.78 & 30.27 & \textbf{0.5835}& 31.61 & 14.24 & 29.63 & 0.7680 & \textbf{44.16} & 24.28 & \textbf{30.23} &\textbf{ 0.7407} \\
    
    ~ & NCLS+MS  & 34.84 & 16.05 & 29.47 & 0.5807 & 31.33 & 13.86 & 29.31 & 0.7675 & 42.68 & 23.51 & 29.24 & 0.7361  \\
    ~ & MCLAS & \textbf{35.65} & \textbf{16.97} & \textbf{31.14} & 0.5770 & \textbf{36.48} & \textbf{17.21} & \textbf{34.86} & \textbf{0.7897} & 42.27 & \textbf{24.60} & 30.09 &  0.7069 \\
    
    \bottomrule
    \end{tabular}
    }
    \caption{F1 scores of ROUGE and BERTScore in Zh2EnSum, En2DeSum and En2ZhSum dataset. R-1, R-2, and R-L represents ROUGE-1, ROUGE-2, and ROUGE-L, respectively. }
    \label{tab:ROUGE&BERTScore}
\end{table*}

\subsection{Training and Inference}
We use multilingual BERT (mBERT) \citep{devlin2019bert} to initialize our Transformer encoder. The decoder is a Transformer decoder with 6 layers. Each attention module has 8 different attention heads. The hidden size of the decoder's self-attention is 768 and that of the feed-forward network is 2048. The final model contains 296,046,231 parameters.
Because the encoder is pretrained when the decoder is randomly initialized, we use two separate optimizers for the encoder and the decoder \citep{liu2019text}. 
The encoder's learning rate $\eta_{e}$ is set as 0.005, while the decoder's learning rate $\eta_d$ is 0.2. 
Warmup-steps for the encoder are 10,000 and 5,000 for the decoder.
We train the model on two TITAN RTX GPUs for one day with gradient accumulation every 5 steps.
Dropout with a probability 0.1 is applied before all the linear layers. 
We find that the target vocabulary type doesn't have much influence on the final result. Therefore, we directly use mBERT's subwords vocabulary as our target vocabulary. 
Nevertheless, in case tokens would be produced in the wrong language, we constructe a target token vocabulary for each target language. In the inference period, we only generate tokens from the corresponding vocabulary. 
During the decoding stage, we use beam search (size 5) and trigram block to avoid repetition. Length penalty is set between 0.6 and 1. All the hyperparameters are manually tuned using PPL and accuracy metric on the validation set.

\subsection{Baselines}
\label{sec:baselines}
We compare MCLAS in low-resource scenarios with the following baselines: 


\paragraph{NCLS} CLS model proposed by \citet{zhu2019ncls}. In low-resource scenarios, we initialize our model with the pretrained MS model and then use a few samples to optimize Equation~(\ref{equ:nclsloss}).
\paragraph{NCLS+MS} Multi-task framework proposed by \citet{zhu2019ncls}. We find that NCLS+MS fails to converge when it is partly initialized by the pretrained MS model (the CLS decoder is randomly initialized). Hence, we fully initialize the multi-task model using the pretrained MS model. Specifically, the two separate decoders are both initialized by the pretrained monolingual decoder. Then the model is optimized with Equation~(\ref{equ:multitaskloss}).
\paragraph{TLTran} Transformer-based Late Translation is a pipeline method. First, a monolingual summarization model summarizes the source document. A translation model is then applied to translate the summary. The summarization model is trained with monolingual document-summary pairs in three datasets. Specifically, we continue using WMT’19 English-German winner as the translation model for En2DeSum. 

Some recent proposed models improve the performance of CLS task. Methods \textbf{NCLS+MT}, \textbf{TETran}~\citep{zhu2019ncls}, and the system proposed by \citet{ouyang-etal-2019-robust}  require external long document machine translation (MT) corpora. 
The method proposed by \citet{cao2020jointly} requires not only parallel summaries but also document pairs translated by MT systems. Another method proposed by \citet{zhu2020attend} requires bilingual lexicons extracted from large parallel MT datasets (2.08M sentence pairs from eight LDC
corpora).  We choose not to use these models as baselines since comparing MCLAS with them is unfair.

\subsection{Automatic Evaluation Results}
The overall results under low-resource scenarios and full-dataset scenario are shown in Table~\ref{tab:ROUGE&BERTScore}. 
We reimplement a variety of models and evaluate them using F1 scores of the standard ROUGE metric \citep{lin2004rouge} (ROUGE-1, ROUGE-2, and ROUGE-L) and BERTScore\footnote{https://github.com/Tiiiger/bert\_score}~\citep{zhang2019bertscore}.
The following analysis is from our observations.

In the Zh2EnSum and En2DeSum datasets, MCLAS achieves significant improvements over baselines in all the low-resource scenarios. 
It is worth noting that combining NCLS+MS in our experiments does not bring much improvement to the NCLS model. 
We consider that this is because mBERT has already provided multilingual encoding for our models. 
%

However, we find that in the En2ZhSum dataset, MCLAS did not perform as  well as that in the other two datasets. We speculate that is due to the imbalance of English reference and Chinese reference. The average length of $S^A$ and $S^B$ in En2ZhSum is 55.21 and 95.96, respectively \citep{zhu2019ncls}. This condition largely breaks the alignment between languages, leading to MCLAS the performing slightly weaker.
Despite this, results in En2DeSum and Zh2EnSum demonstrate that our proposed MCLAS model is effective for CLS under limited resources.

Finally, our proposed model also has superior performance compared to baseline models given the full training dataset, achieving the best ROUGE score in En2DeSum and Zh2EnSum datasets.
\begin{table}[t]
\setlength{\tabcolsep}{3pt}
\small
  \centering
  \resizebox{0.99\linewidth}{!}{
    \begin{tabular}{lrrrrrrrrr}
    \toprule
    \multirow{2}*{Models} & \multicolumn{3}{c}{Minimum} & \multicolumn{3}{c}{Medium}  & \multicolumn{3}{c}{Maximum} \\
    \cmidrule(r{4pt}){2-4} \cmidrule(r{4pt}){5-7} \cmidrule(l){8-10}
    ~ & IF & CC & FL & IF & CC & FL & IF & CC & FL \\
    \midrule
    MCLAS &\textbf{-0.264} &\textbf{0.164} &\textbf{-0.021} & 0.000 & \textbf{0.236} & \textbf{0.164} &\textbf{0.057} &\textbf{0.464} & \textbf{0.214} \\
    NCLS & -0.243 & -0.386 & -0.364 & \textbf{0.036} & -0.221 & -0.257 & -0.129 & -0.329 & -0.186\\ 
    NCLS+MS & -0.371 & -0.407 & -0.286 & -0.343 & -0.536 & -0.407 & -0.179 & -0.364 & -0.214\\
    GOLD & 0.879 & 0.629 & 0.671 & 0.300 & 0.529 & 0.500 & 0.257 & 0.221 & 0.179\\
    \bottomrule
    \end{tabular}
    
    }
\caption{Human evaluation results in Zh2EnSum dataset. The best results are in bold.}
    \label{tab:human_eval}
\end{table}
\begin{table}[t]
\small
  \centering
  \resizebox{0.99\linewidth}{!}{

        \begin{tabular}{lrr}
    \toprule
    Scenarios & Fleiss' Kappa & Overall Agreement \\
    \midrule
    Minimum & 0.37 & 60.48\% \\
    Medium & 0.22 & 51.35\% \\
    Maxmium & 0.20 & 50.16\% \\
    \bottomrule 
    \end{tabular}
        }
    \caption{Fleiss' Kappa and overall agreement percent of our human evaluation results. A higher value indicates higher agreements among participants.}
    \label{tab:inter_agreement}
\end{table}

\begin{table}[t]
\small
  \centering
  \resizebox{0.99\linewidth}{!}{
    \begin{tabular}{llrr}
    \toprule
    Scenarios &{Models} & En2DeSum & Zh2EnSum  \\
    \midrule
    \multirow{3}*{\makecell[l]{Minimum\\ Low-resource\\ Scenario}}  & NCLS & 13.48 ($+$4.69) & 18.49 ($+$3.51)  \\
    
    ~ & NCLS+MS  & 12.83 ($+$4.04)  & 18.68 ($+$3.70)  \\
    ~ & MCLAS & 7.80 \textbf{($-$0.90)} & 13.16 \textbf{($-$1.82)}  \\
    \midrule
    \multirow{3}*{\makecell[l]{Medium\\ Low-resource\\ Scenario}}  & NCLS &13.13 ($+$4.34)& 18.60 ($+$3.62)  \\
    
    ~ & NCLS+MS  & 12.90 ($+$4.11) &18.57 ($+$3.59) \\
    ~ & MCLAS & 8.65 \textbf{($-$0.14)} & 13.10 \textbf{($-$1.88)}  \\
    \midrule
    \multirow{3}*{\makecell[l]{Maximum\\ Low-resource\\ Scenario}}  & NCLS & 13.37 ($+$4.58) & 18.44 ($+$3.46)  \\
    
    ~ & NCLS+MS  & 13.37 ($+$4.58) & 18.75 ($+$3.77)  \\
    ~ & MCLAS & 8.46 \textbf{($-$0.33)} &12.83 \textbf{($-$2.15)}  \\
    \midrule
    \multicolumn{2}{c}{Gold}  & 8.79 & 14.98  \\
    \bottomrule
    \end{tabular}
    
    }
\caption{Target summary length generated by various models. The best results are in bold.}
    \label{tab:target_length}
\end{table}

\begin{table}[t]
\small
  \centering
  \resizebox{0.99\linewidth}{!}{
    \begin{tabular}{lcccccc}
    \toprule
    \multirow{2}*{Models} & \multicolumn{3}{c}{En2DeSum} & \multicolumn{3}{c}{Zh2EnSum}\\
    \cmidrule(r{4pt}){2-4} \cmidrule(l){5-7}
     ~ & R-1 & R-2 & R-L & R-1 & R-2 & R-L\\
     \midrule
     MS-Pretrain & 39.16 & 19.21 & 35.42 & 41.71 &\textbf{27.76} & \textbf{37.97} \\
     NCLS+MS & 36.06 & 16.84 & 32.72 & 39.98 & 26.03 & 36.13 \\
     MCLAS &\textbf{ 45.59 }& \textbf{23.77} & \textbf{42.51} & \textbf{41.72} & 27.69 & 37.92 \\
     \bottomrule
    \end{tabular}
    }
\caption{Monolingual summary results in Zh2EnSum and En2ZhSum datasets. MS-Pretrain refers to the pretrained model for monolingual summarization. }
    \label{tab:mono}
\end{table}
\begin{table}[t]
\small
  \centering
  \resizebox{0.99\linewidth}{!}{
    \begin{tabular}{lcccc}
    \toprule
    Metrics & MS-Pretrain&NCLS+MS&MCLAS&Ground Truth\\
    \midrule
    R-1 Recall	&\textbf{58.37}	&52.62	&46.88	&-\\
R-1 Precision	&30.45	&28.41	&\textbf{46.31}	&-\\ 
    \midrule
R-2 Recall	&\textbf{30.21}	&25.91	&24.54	&- \\
R-2 Precision	&14.65	&12.99	&\textbf{24.11}	&- \\
    \midrule
R-L Recall	&\textbf{52.97}	&47.89	&43.74	&- \\
R-L Precision	&27.51	&25.71	&\textbf{43.15}	&- \\
    \midrule
Avg. Length	&18.64 (+9.53)	&17.98 (+8.87)	& 9.15 \textbf{(+0.04)}	&9.11 \\
    \bottomrule 
    \end{tabular}
    
        }
\caption{Analysis on monolingual summary generation ability of MCLAS trained with En2DeSum dataset.}
    \label{tab:en2demono}
\end{table}
\subsection{Human Evaluation}
In addition to automatic evaluation, we conduct a human evaluation to verify our model's performance. We randomly chose 60 examples (20 for each low-resource scenario) from the Zh2EnSum test dataset. Seven graduate students with high levels of fluency in English and Chinese are asked to assess the generated summaries and gold summaries from independent perspectives: \textbf{informativeness}, \textbf{fluency}, and \textbf{conciseness}. 
We follow the Best-Worst Scaling method \citep{kiritchenko2017best}. Participants are asked to indicate the best and worst items from each perspective. The result scores are calculated based on the percentage of times each system is selected as best minus the times it is selected as worst. 
Hence, final scores range from -1 (worst) to 1 (best).
Results are shown in Table~\ref{tab:human_eval}.

As the data size increases, all the models achieve better results. Our proposed MCLAS outperformed NCLS and NCLS+MS in all the metrics. 
We notice that MCLAS is especially strong in conciseness. This phenomenon will be analyzed in Section \ref{subsec:summarylength}

We show Fleiss' Kappa scores of our conducted human evaluation in Table~\ref{tab:inter_agreement}, which demonstrates a good inter-agreement among the participants.

\begin{figure}[t]
\centering
\includegraphics[width=0.99\columnwidth]{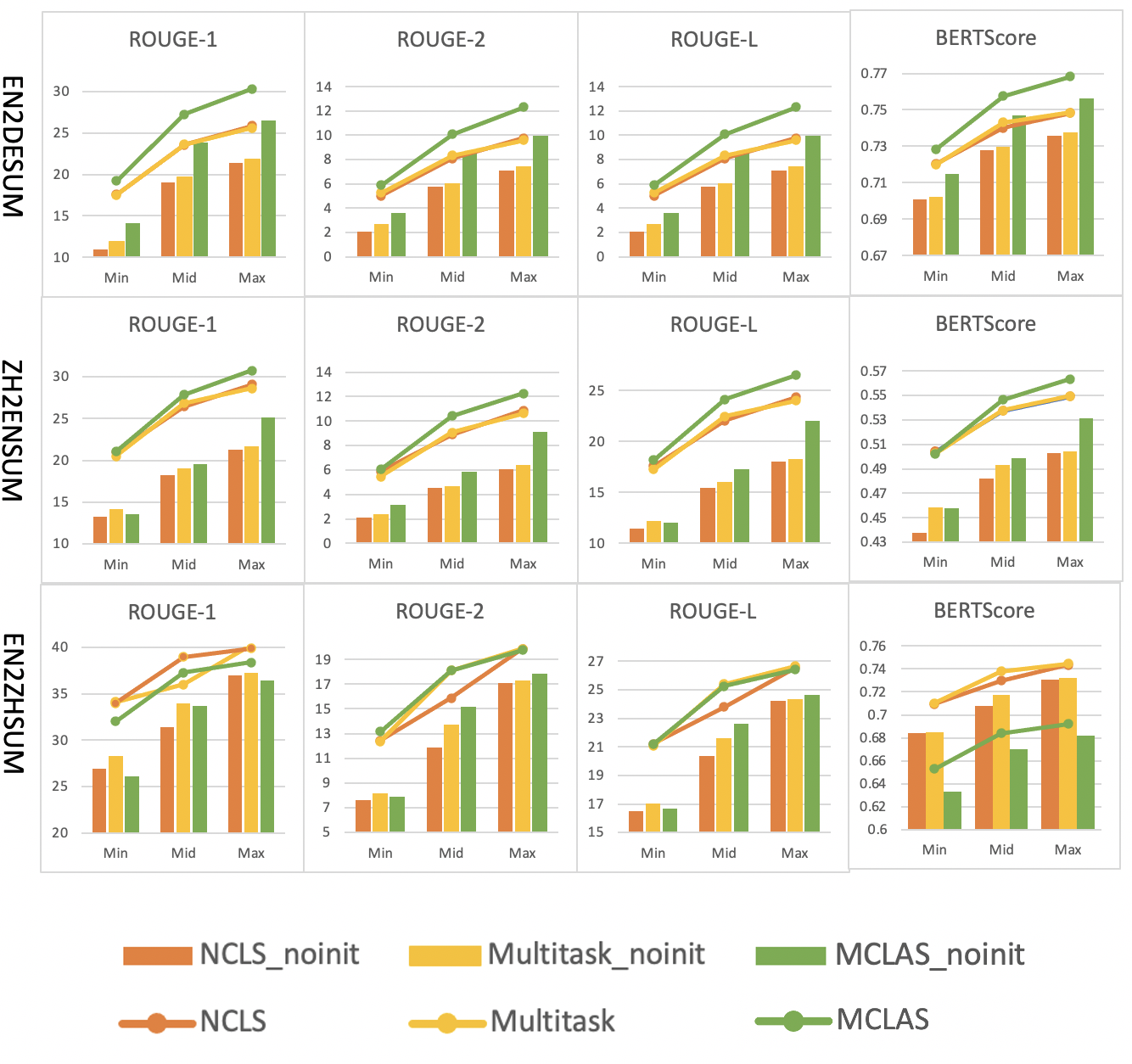} 
\caption{Line and column chart of En2DeSum and Zh2EnSum results. Lines represent models initialized with pretrained monolingual summarization model. Columns represent models trained from scratch.}
\label{fig:line_chart}
\end{figure}

\subsection{Analysis on Initialization Methods}
We use a monolingual summarization model to initialize our model. However, whether this initialization method works is still in question. Therefore we compare our models with non-initialized models, shown in Figure~\ref{fig:line_chart}. Among the three datasets, the initialization methods bring a huge improvement to all of the models.

\subsection{Analysis on Summary Length}
\label{subsec:summarylength}
One of the goals of automatic summarization is to produce brief text. Yet many neural auto-regressive models tend to produce a longer summary to improve the recall metric. 
Results in Table~\ref{tab:target_length} show that interactions enable MCLAS to generate shorter summaries than other models, which more closely resembles human summaries.
We can safely conclude that MCLAS can keep the summary in a fairly appropriate length, leading to concise generated summaries.
We speculate that this is due to its ability to capture interactions between languages, conditioning cross-lingual summaries on relatively precise monolingual summaries.

\subsection{Analysis on Monolingual Summarization}
Modeling interactions between languages brings many advantages. Specifically, we find that MCLAS can preserve more monolingual summarization knowledge than the NCLS+MS model during low-resource fine-tuning, or even promote its performance.
We generate monolingual summaries with models trained in the maximum low-resource scenario. 
In Table~\ref{tab:mono}, we can clearly see that MCLAS retains more monolingual summarization knowledge in the Zh2EnSum dataset. In the En2DeSum dataset, monolingual summarization performance is even significantly improved. We speculate that this is due to MCLAS's ability to provide the interactions between languages.
\begin{figure}[t]
\centering
\includegraphics[width=0.99\columnwidth]{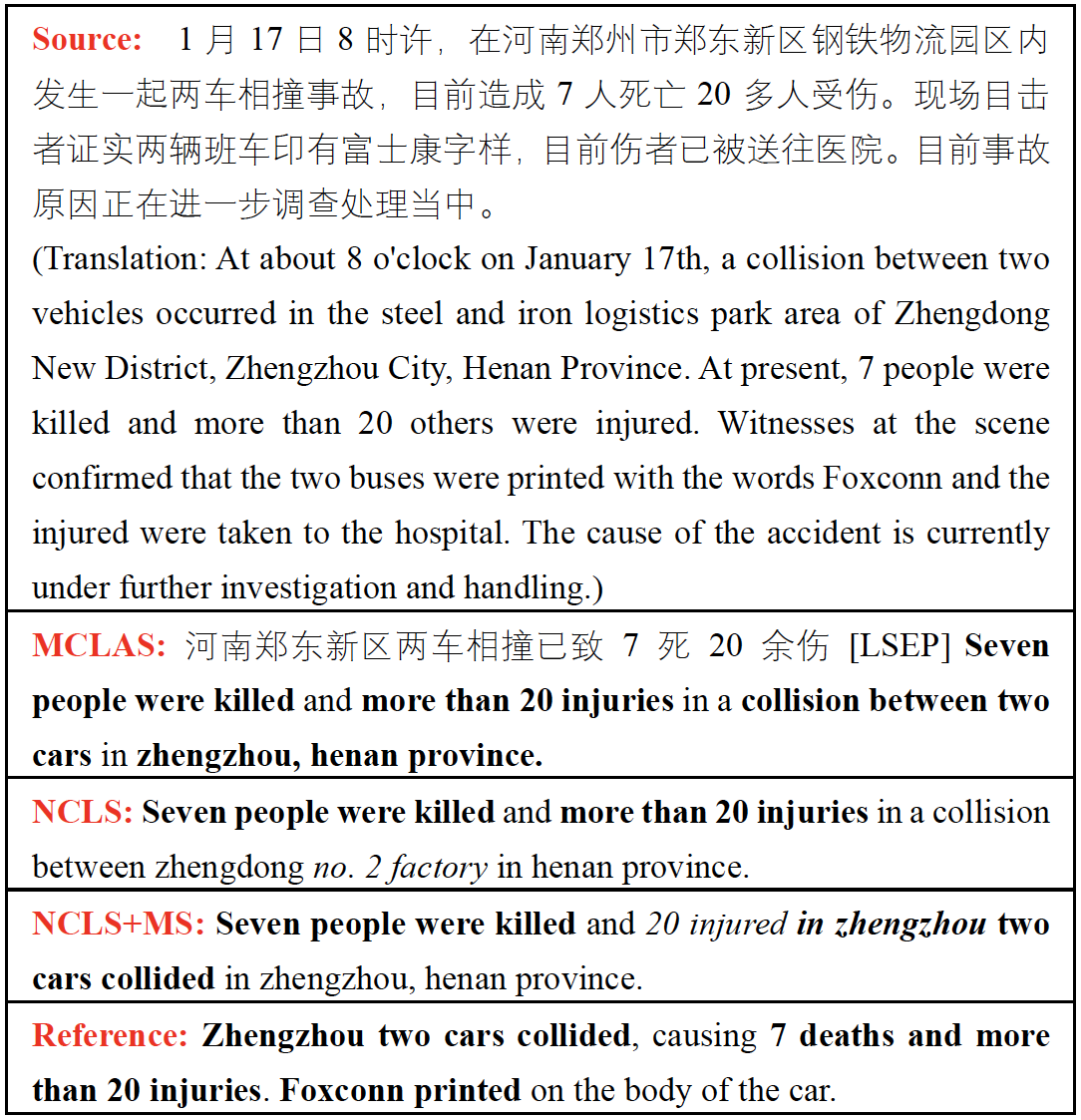} 
\caption{An example of generated cross-lingual summaries. Important phrases are bold while incorrect information generated by each model is italicized. Non-fluent parts in sentences are bold and italicized. }
\label{casestudy}
\end{figure}

We focus specifically on digging into results in En2DeSum, evaluating its detailed ROUGE and average summary length, presented in Table~\ref{tab:en2demono}. 
We find that ROUGE improvement mainly resulted from precision while recall barely decrease the performances. This and the Avg. length metric shows that MCLAS produces more precise summaries while retaining most of the important information, leading to the metric increase in ROUGE.


\subsection{Case Study}
\label{sec:casestudy}

In Figure~\ref{casestudy}, on the Zh2EnSum dataset, there is a list comparing the reference summary and outputs of models trained in the maximum low-resource scenario.  
Clearly, the NCLS model loses the information ``two cars'' and generates the wrong information ``No.2 factory''. The NCLS+MS model is not accurate when describing the number of injured people, dropping important information ``more than''. Additionally, the NCLS+MS model also has fluency and repetition issues: ``in zhengzhou'' appears twice in its generated summary.  In contrast, MCLAS captures all of this information mentioned in both its Chinese and English output, and the English summary is well aligned with the Chinese summary. Finally, all of the models ignore the information ``foxconn printed on the body of the car''. See Appendix \ref{appendix:samples} for more examples.
\begin{figure}[t]
\centering
\includegraphics[width=0.99\columnwidth]{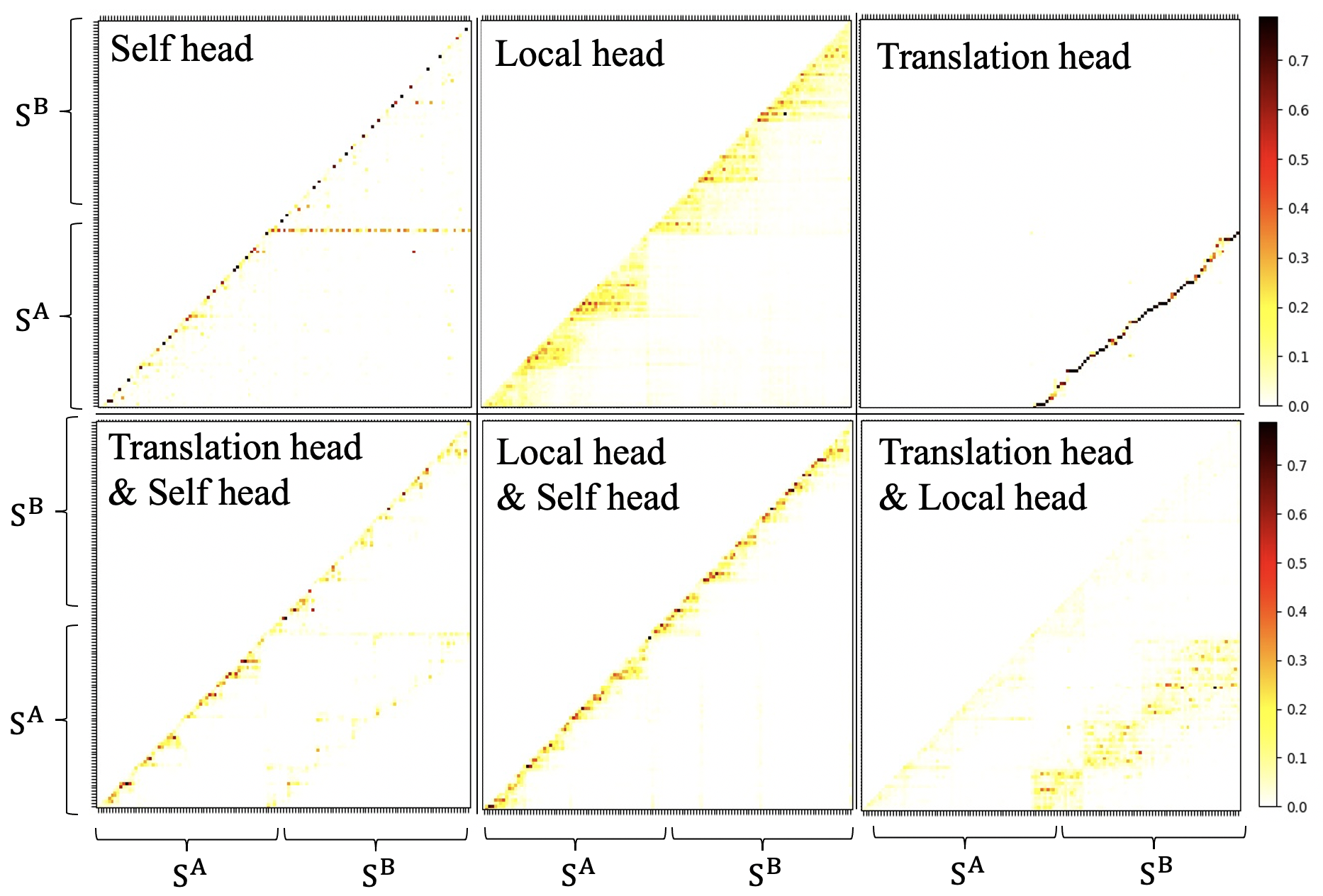} 
\caption{Different types of self-attention heads in MCLAS's decoder. The x-axis and y-axis are both concatenated source-language summary $S^A$ and target-language summary $S^B$ tokens. 
Darker color shows the more highly related associations between tokens. The horizontal line in self head represents the \textsc{[lsep]} token. Some attention heads attend to \textsc{[lsep]} to confirm whether it is generating a cross-lingual summary. }
\label{self}
\end{figure}

\begin{figure}[t]
\centering
\includegraphics[width=0.99\columnwidth]{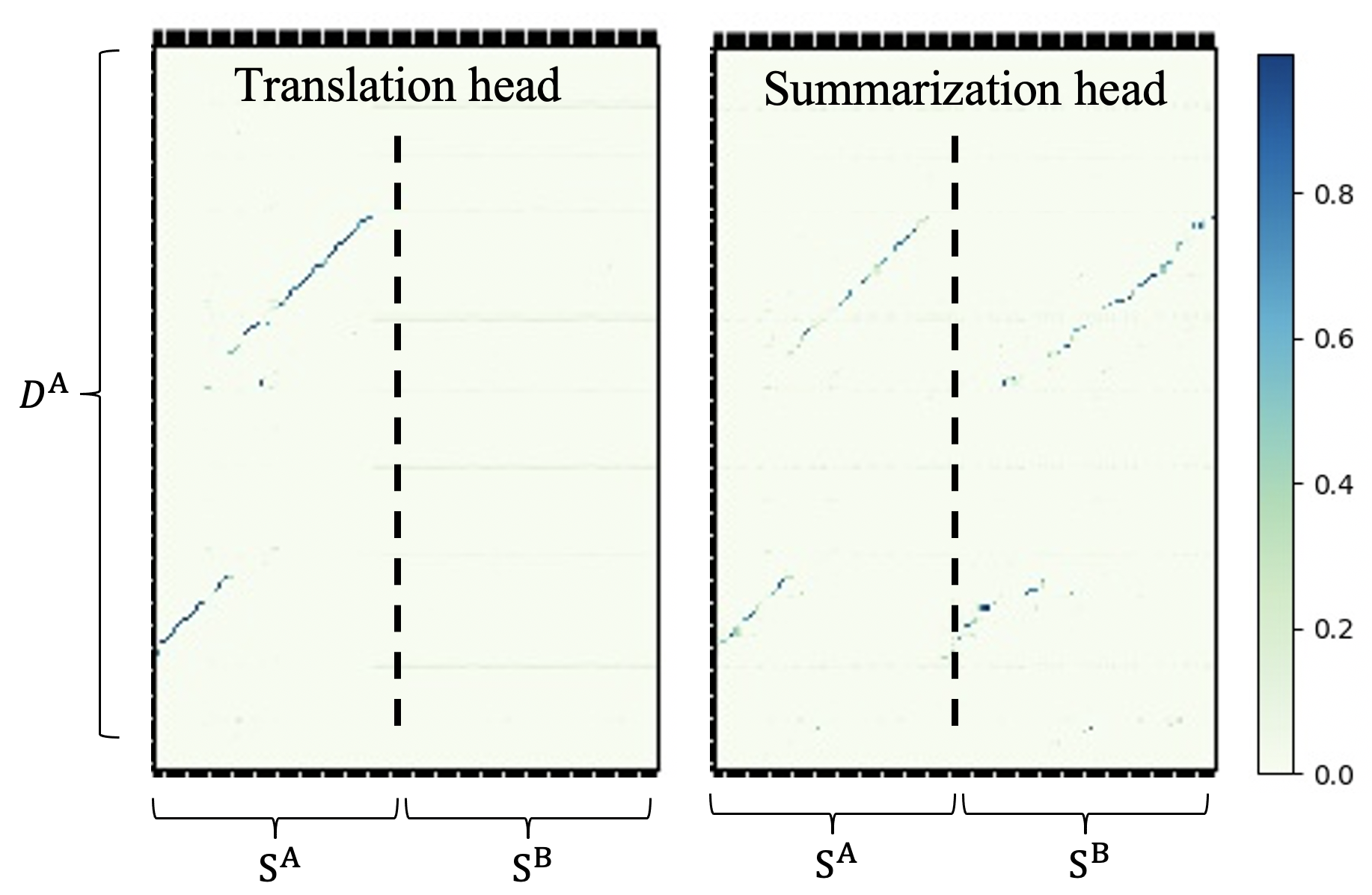} 
\caption{Different types of encoder-decoder attention heads in MCLAS's decoder. The x-axis represents concatenated source-language summary $S^A$ and target-language summary $S^B$ tokens while the y-axis is the document $D^A$ tokens.
In news texts, important information often gathers in the front part of the document. We only retain the informative part of the y-axis, omitting the blank part that the model do not attend to.}
\label{context}
\end{figure}
\section{Probing into Attention Heads}
\label{sec:probing}
We have observed a successful alignment between $S^A$ and $S^B$ produced by our model in Section~\ref{sec:casestudy}. In this section, we dig into this and analyze how the model learns the relationships. 
For a CLS task from document $D^A$ to $S^B$, our hypotheses are:  (1) the unified decoder is implicitly  undertaking translation from $S^A$ to $S^B$; (2) the unified decoder also conducts both monolingual and cross-lingual summarization. 
To verify these hypotheses, we visualize attention distributions of the Transformer decoders trained on En2ZhSum. Neural models can be explicitly explained using probing into the attention heads~\citep{michel2019sixteen,voita2019analyzing}. We follow the previous work and visualize the function of all attention heads in the decoder to verify the relationships of the concatenated cross-lingual summaries (i.e., translation) and cross-lingual document-summary pairs (i.e., summarization).
%


\subsection{Analysis on Translation}
We assume that the decoder translates only if the source summary $S^A$ and the target summary $S^B$ align well. This means that MCLAS is transferring knowledge from $S^A$ to $S^B$. 
We visualize and probe all 48 \textbf{self-attention heads} in the unified decoder.
We find 23 (47.9\%) \textbf{translation heads}, defined as the heads attending from $y_j^B$ to the corresponding words in language $A$. These heads undertake a translation function. 19 (39.6\%) heads are \textbf{local heads}, attending to a few words before them and modeling context information. 12 (25\%) heads are \textbf{self heads}, which only attend to themselves to retain the primary information. Some of the heads can be categorized into two types. Note that all of the heads behave similarly across different samples. We find that most of the heads are translation heads, indicating that our unified decoder is translating $S^A$ into $S^B$. We sample some representative heads in Figure~\ref{self} to show their functionalities.

\subsection{Analysis on Summarization}
To analyze whether the decoder for $S^B$ is simply translating from $S^A$ or that it also summarizes the source document, we visualize the distribution of 48 \textbf{encoder-decoder attention} heads.
We find 28 (58.3\%) \textbf{summarization heads} that attend to the document's important parts when generating both the monolingual summary and the cross-lingual summary. We also find 20 (41.7\%) \textbf{translation heads}, which focus on the source document when generating $S^A$, while focusing on nothing when generating $S^B$. 
We speculate that summarization heads are responsible for the summarization function and that translation heads cut down the relation between $S^B$ and source document $D^A$, leaving space for translation. 
Again, all the heads behave similarly across different samples.
We select two representative samples in Figure~\ref{context}.

The existence of both summarization and translation heads in encoder-decoder attention components supports our views: the unified decoder simultaneously conducts translation and summarization. Therefore, our model enhances the interactions between different languages, being able to facilitate cross-lingual summarization under low-resource scenarios. See Appendix~\ref{appendix:attn} for detailed visualization results.

\section{Discussions}
An ideal low-resource experiment should be conducted with real low-resource languages. Although possible, it takes much effort to acquire such datasets. 
Hence, it is the second-best choice that we simulate our low-resource scenarios by artificially limiting the amount of the available data. 
Some may question it about the feasibility of our method in real low-resource languages since machine translation systems, which is used to generate document-summary pairs, would be of lower quality for truly low-resource languages. 
For this concern, we consider it still possible to acquire thousands of high-quality human translated parallel summaries, as \citet{duan-etal-2019-zero} adopt on their test set, to apply our method.

\section{Conclusion}
In this paper, we propose a novel multi-task learning framework MCLAS to achieve cross-lingual abstractive summarization with limited parallel resources. Our model shares a unified decoder that sequentially generates both monolingual and cross-lingual summaries. Experiments on two cross-lingual summarization datasets demonstrate that our framework outperforms all the baseline models in low-resource and full-dataset scenarios. 

\section*{Acknowledgements}
This work is supported by the Joint Funds of the
National Natural Science Foundation of China (Grant No. U19B2020), the Funds of the Integrated Application Software Project.
We appreciate the helpful discussions with Sanxing Chen, Jia-Ao Zhan, Xuyang Lu, Xiao Liu, and Yuxiang Zhou. We also thank all the anonymous reviewers for their insightful suggestions.


\bibliographystyle{acl_natbib}
\bibliography{acl2021,anthology}

\begin{thebibliography}{30}
\expandafter\ifx\csname natexlab\endcsname\relax\def\natexlab#1{#1}\fi

\bibitem[{Cao et~al.(2020)Cao, Liu, and Wan}]{cao2020jointly}
Yue Cao, Hui Liu, and Xiaojun Wan. 2020.
\newblock Jointly learning to align and summarize for neural cross-lingual
  summarization.
\newblock In \emph{Proceedings of the 58th Annual Meeting of the Association
  for Computational Linguistics}, pages 6220--6231.

\bibitem[{Chi et~al.(2020)Chi, Dong, Wei, Wang, Mao, and Huang}]{chi2020cross}
Zewen Chi, Li~Dong, Furu Wei, Wenhui Wang, Xian-Ling Mao, and Heyan Huang.
  2020.
\newblock Cross-lingual natural language generation via pre-training.
\newblock In \emph{AAAI}, pages 7570--7577.

\bibitem[{Devlin et~al.(2019)Devlin, Chang, Lee, and
  Toutanova}]{devlin2019bert}
Jacob Devlin, Ming-Wei Chang, Kenton Lee, and Kristina Toutanova. 2019.
\newblock Bert: Pre-training of deep bidirectional transformers for language
  understanding.
\newblock In \emph{Proceedings of the 2019 Conference of the North American
  Chapter of the Association for Computational Linguistics: Human Language
  Technologies, Volume 1 (Long and Short Papers)}, pages 4171--4186.

\bibitem[{Dong et~al.(2019)Dong, Yang, Wang, Wei, Liu, Wang, Gao, Zhou, and
  Hon}]{NEURIPS2019_c20bb2d9}
Li~Dong, Nan Yang, Wenhui Wang, Furu Wei, Xiaodong Liu, Yu~Wang, Jianfeng Gao,
  Ming Zhou, and Hsiao-Wuen Hon. 2019.
\newblock \href
  {https://proceedings.neurips.cc/paper/2019/file/c20bb2d9a50d5ac1f713f8b34d9aac5a-Paper.pdf}
  {Unified language model pre-training for natural language understanding and
  generation}.
\newblock In \emph{Advances in Neural Information Processing Systems},
  volume~32. Curran Associates, Inc.

\bibitem[{Duan et~al.(2019{\natexlab{a}})Duan, Yin, Zhang, Chen, and
  Luo}]{duan2019zero}
Xiangyu Duan, Mingming Yin, Min Zhang, Boxing Chen, and Weihua Luo.
  2019{\natexlab{a}}.
\newblock Zero-shot cross-lingual abstractive sentence summarization through
  teaching generation and attention.
\newblock In \emph{Proceedings of the 57th Annual Meeting of the Association
  for Computational Linguistics}, pages 3162--3172.

\bibitem[{Duan et~al.(2019{\natexlab{b}})Duan, Yin, Zhang, Chen, and
  Luo}]{duan-etal-2019-zero}
Xiangyu Duan, Mingming Yin, Min Zhang, Boxing Chen, and Weihua Luo.
  2019{\natexlab{b}}.
\newblock \href {https://doi.org/10.18653/v1/P19-1305} {Zero-shot cross-lingual
  abstractive sentence summarization through teaching generation and
  attention}.
\newblock In \emph{Proceedings of the 57th Annual Meeting of the Association
  for Computational Linguistics}, pages 3162--3172, Florence, Italy.
  Association for Computational Linguistics.

\bibitem[{Gu et~al.(2018)Gu, Wang, Chen, Li, and Cho}]{gu2018meta}
Jiatao Gu, Yong Wang, Yun Chen, Victor~OK Li, and Kyunghyun Cho. 2018.
\newblock Meta-learning for low-resource neural machine translation.
\newblock In \emph{Proceedings of the 2018 Conference on Empirical Methods in
  Natural Language Processing}, pages 3622--3631.

\bibitem[{Hermann et~al.(2015)Hermann, Kocisky, Grefenstette, Espeholt, Kay,
  Suleyman, and Blunsom}]{hermann2015teaching}
Karl~Moritz Hermann, Tomas Kocisky, Edward Grefenstette, Lasse Espeholt, Will
  Kay, Mustafa Suleyman, and Phil Blunsom. 2015.
\newblock Teaching machines to read and comprehend.
\newblock In \emph{Advances in neural information processing systems}, pages
  1693--1701.

\bibitem[{Kiritchenko and Mohammad(2017)}]{kiritchenko2017best}
Svetlana Kiritchenko and Saif Mohammad. 2017.
\newblock Best-worst scaling more reliable than rating scales: A case study on
  sentiment intensity annotation.
\newblock In \emph{Proceedings of the 55th Annual Meeting of the Association
  for Computational Linguistics (Volume 2: Short Papers)}, pages 465--470.

\bibitem[{Lin(2004)}]{lin2004rouge}
Chin-Yew Lin. 2004.
\newblock Rouge: A package for automatic evaluation of summaries.
\newblock \emph{Text Summarization Branches Out}.

\bibitem[{Liu and Lapata(2019)}]{liu2019text}
Yang Liu and Mirella Lapata. 2019.
\newblock Text summarization with pretrained encoders.
\newblock In \emph{Proceedings of the 2019 Conference on Empirical Methods in
  Natural Language Processing and the 9th International Joint Conference on
  Natural Language Processing (EMNLP-IJCNLP)}, pages 3721--3731.

\bibitem[{Michel et~al.(2019)Michel, Levy, and Neubig}]{michel2019sixteen}
Paul Michel, Omer Levy, and Graham Neubig. 2019.
\newblock Are sixteen heads really better than one?
\newblock In \emph{Advances in Neural Information Processing Systems}, pages
  14014--14024.

\bibitem[{Motiian et~al.(2017)Motiian, Jones, Iranmanesh, and
  Doretto}]{motiian2017few}
Saeid Motiian, Quinn Jones, Seyed~Mehdi Iranmanesh, and Gianfranco Doretto.
  2017.
\newblock Few-shot adversarial domain adaptation.
\newblock In \emph{NIPS}.

\bibitem[{Ouyang et~al.(2019)Ouyang, Song, and
  McKeown}]{ouyang-etal-2019-robust}
Jessica Ouyang, Boya Song, and Kathy McKeown. 2019.
\newblock \href {https://doi.org/10.18653/v1/N19-1204} {A robust abstractive
  system for cross-lingual summarization}.
\newblock In \emph{Proceedings of the 2019 Conference of the North {A}merican
  Chapter of the Association for Computational Linguistics: Human Language
  Technologies, Volume 1 (Long and Short Papers)}, pages 2025--2031,
  Minneapolis, Minnesota. Association for Computational Linguistics.

\bibitem[{Radford et~al.(2019)Radford, Wu, Child, Luan, Amodei, and
  Sutskever}]{radford2019language}
Alec Radford, Jeffrey Wu, Rewon Child, David Luan, Dario Amodei, and Ilya
  Sutskever. 2019.
\newblock Language models are unsupervised multitask learners.
\newblock \emph{OpenAI Blog}, 1(8):9.

\bibitem[{Shen et~al.(2018)Shen, Chen, Yang, Liu, Sun et~al.}]{shen2018zero}
Shi-qi Shen, Yun Chen, Cheng Yang, Zhi-yuan Liu, Mao-song Sun, et~al. 2018.
\newblock Zero-shot cross-lingual neural headline generation.
\newblock \emph{IEEE/ACM Transactions on Audio, Speech, and Language
  Processing}, 26(12):2319--2327.

\bibitem[{Song et~al.(2019)Song, Tan, Qin, Lu, and Liu}]{song2019mass}
Kaitao Song, Xu~Tan, Tao Qin, Jianfeng Lu, and Tie-Yan Liu. 2019.
\newblock Mass: Masked sequence to sequence pre-training for language
  generation.
\newblock In \emph{International Conference on Machine Learning}, pages
  5926--5936.

\bibitem[{Vaswani et~al.(2017)Vaswani, Shazeer, Parmar, Uszkoreit, Jones,
  Gomez, Kaiser, and Polosukhin}]{vaswani2017attention}
Ashish Vaswani, Noam Shazeer, Niki Parmar, Jakob Uszkoreit, Llion Jones,
  Aidan~N Gomez, {\L}ukasz Kaiser, and Illia Polosukhin. 2017.
\newblock Attention is all you need.
\newblock In \emph{Advances in neural information processing systems}, pages
  5998--6008.

\bibitem[{Voita et~al.(2019)Voita, Talbot, Moiseev, Sennrich, and
  Titov}]{voita2019analyzing}
Elena Voita, David Talbot, Fedor Moiseev, Rico Sennrich, and Ivan Titov. 2019.
\newblock Analyzing multi-head self-attention: Specialized heads do the heavy
  lifting, the rest can be pruned.
\newblock In \emph{Proceedings of the 57th Annual Meeting of the Association
  for Computational Linguistics}, pages 5797--5808.

\bibitem[{Wan(2011)}]{wan2011using}
Xiaojun Wan. 2011.
\newblock Using bilingual information for cross-language document
  summarization.
\newblock In \emph{Proceedings of the 49th Annual Meeting of the Association
  for Computational Linguistics: Human Language Technologies}, pages
  1546--1555.

\bibitem[{Wan et~al.(2010)Wan, Li, and Xiao}]{wan2010cross}
Xiaojun Wan, Huiying Li, and Jianguo Xiao. 2010.
\newblock Cross-language document summarization based on machine translation
  quality prediction.
\newblock In \emph{Proceedings of the 48th Annual Meeting of the Association
  for Computational Linguistics}, pages 917--926.

\bibitem[{Wang et~al.(2020)Wang, Yao, Kwok, and Ni}]{wang2020generalizing}
Yaqing Wang, Quanming Yao, James~T Kwok, and Lionel~M Ni. 2020.
\newblock Generalizing from a few examples: A survey on few-shot learning.
\newblock \emph{ACM Computing Surveys (CSUR)}, 53(3):1--34.

\bibitem[{Yan et~al.(2015)Yan, Yap, and Mori}]{yan2015multi}
Wang Yan, Jordan Yap, and Greg Mori. 2015.
\newblock Multi-task transfer methods to improve one-shot learning for
  multimedia event detection.
\newblock In \emph{BMVC}, pages 37--1.

\bibitem[{Yao et~al.(2015)Yao, Wan, and Xiao}]{yao2015phrase}
Jin-ge Yao, Xiaojun Wan, and Jianguo Xiao. 2015.
\newblock Phrase-based compressive cross-language summarization.
\newblock In \emph{Proceedings of the 2015 conference on empirical methods in
  natural language processing}, pages 118--127.

\bibitem[{Zhang et~al.(2016)Zhang, Zhou, and Zong}]{zhang2016abstractive}
Jiajun Zhang, Yu~Zhou, and Chengqing Zong. 2016.
\newblock Abstractive cross-language summarization via translation model
  enhanced predicate argument structure fusing.
\newblock \emph{IEEE/ACM Transactions on Audio, Speech, and Language
  Processing}, 24(10):1842--1853.

\bibitem[{Zhang et~al.(2019{\natexlab{a}})Zhang, Zhao, Saleh, and
  Liu}]{zhang2019pegasus}
Jingqing Zhang, Yao Zhao, Mohammad Saleh, and Peter~J Liu. 2019{\natexlab{a}}.
\newblock Pegasus: Pre-training with extracted gap-sentences for abstractive
  summarization.
\newblock \emph{arXiv preprint arXiv:1912.08777}.

\bibitem[{Zhang et~al.(2019{\natexlab{b}})Zhang, Kishore, Wu, Weinberger, and
  Artzi}]{zhang2019bertscore}
Tianyi Zhang, Varsha Kishore, Felix Wu, Kilian~Q Weinberger, and Yoav Artzi.
  2019{\natexlab{b}}.
\newblock Bertscore: Evaluating text generation with bert.
\newblock In \emph{International Conference on Learning Representations}.

\bibitem[{Zhu et~al.(2018)Zhu, Li, Liu, Zhou, Zhang, and Zong}]{zhu2018msmo}
Junnan Zhu, Haoran Li, Tianshang Liu, Yu~Zhou, Jiajun Zhang, and Chengqing
  Zong. 2018.
\newblock Msmo: Multimodal summarization with multimodal output.
\newblock In \emph{Proceedings of the 2018 conference on empirical methods in
  natural language processing}, pages 4154--4164.

\bibitem[{Zhu et~al.(2019)Zhu, Wang, Wang, Zhou, Zhang, Wang, and
  Zong}]{zhu2019ncls}
Junnan Zhu, Qian Wang, Yining Wang, Yu~Zhou, Jiajun Zhang, Shaonan Wang, and
  Chengqing Zong. 2019.
\newblock Ncls: Neural cross-lingual summarization.
\newblock In \emph{Proceedings of the 2019 Conference on Empirical Methods in
  Natural Language Processing and the 9th International Joint Conference on
  Natural Language Processing (EMNLP-IJCNLP)}, pages 3045--3055.

\bibitem[{Zhu et~al.(2020)Zhu, Zhou, Zhang, and Zong}]{zhu2020attend}
Junnan Zhu, Yu~Zhou, Jiajun Zhang, and Chengqing Zong. 2020.
\newblock Attend, translate and summarize: An efficient method for neural
  cross-lingual summarization.
\newblock In \emph{Proceedings of the 58th Annual Meeting of the Association
  for Computational Linguistics}, pages 1309--1321.

\end{thebibliography}

\clearpage
\appendix

\section{Samples}
\label{appendix:samples}
\begin{figure}[!htb]
\centering
\includegraphics[width=0.96\columnwidth]{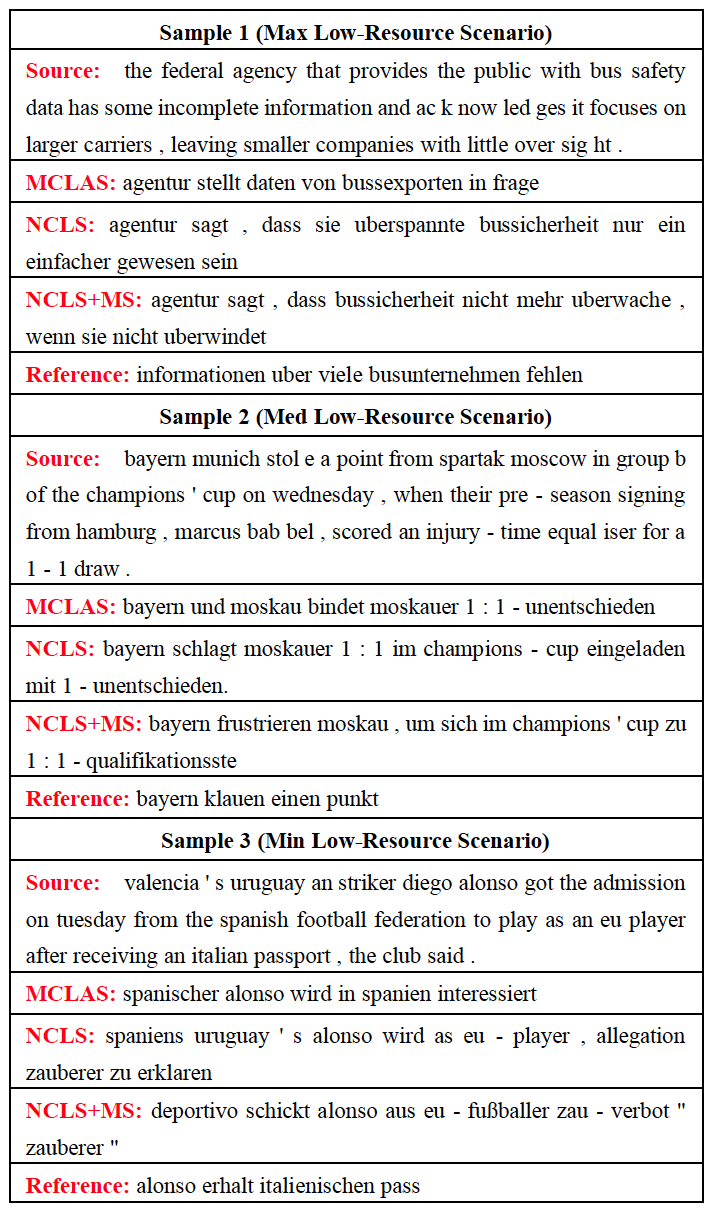} 
\caption{Examples of models trained in En2DeSum dataset. }
\label{caseen2de}
\end{figure}

\begin{figure*}[t]
\centering
\includegraphics[width=1.96\columnwidth]{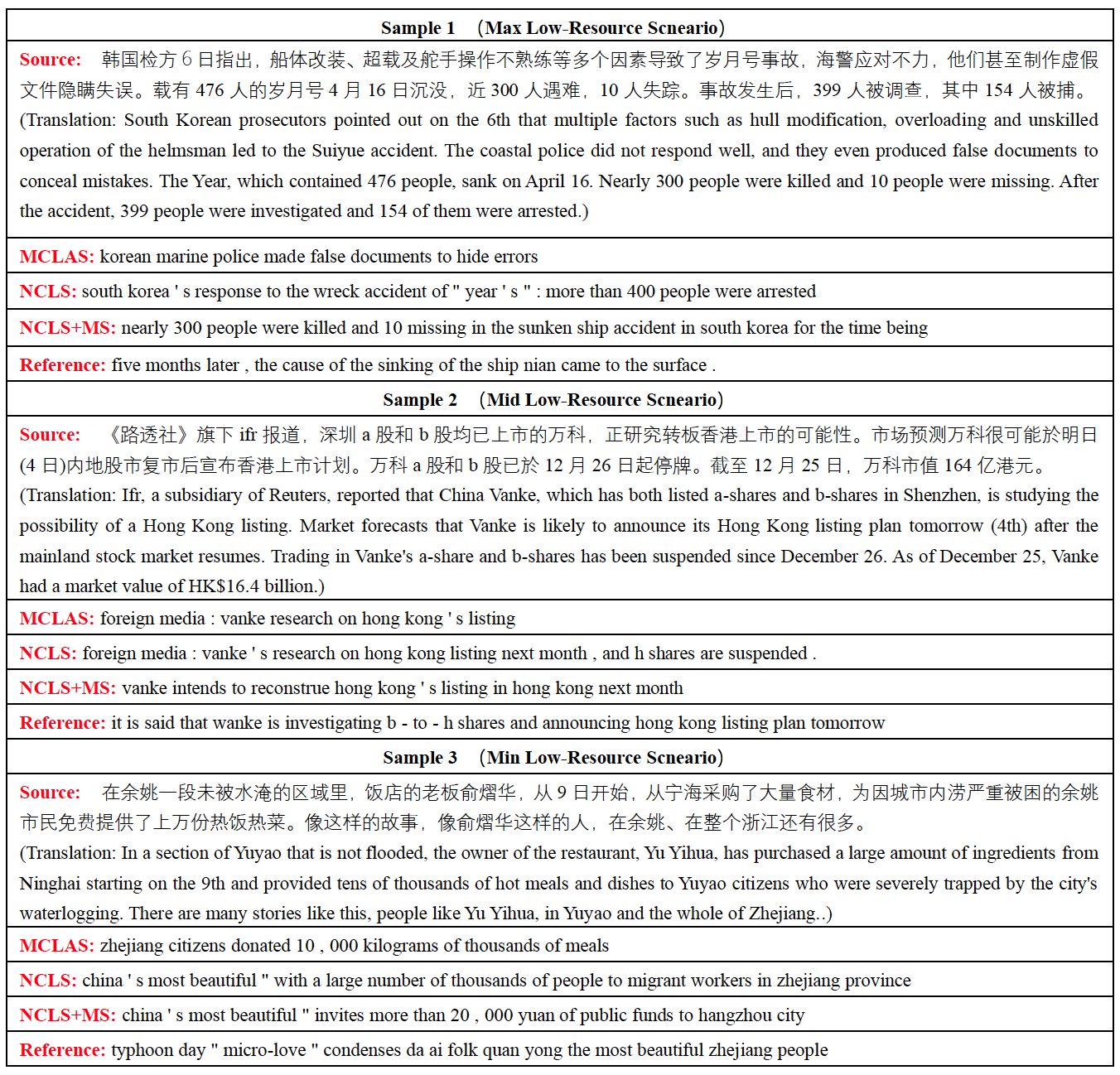} 
\caption{Examples of models trained in Zh2EnSum dataset }
\label{caseen2zh}
\end{figure*}

We list some samples from outputs of various models. Samples from En2DeSum dataset are shown in Figure~\ref{caseen2de}. Samples from Zh2EnSum dataset are shown in Figure~\ref{caseen2zh}. We randomly selected one sample from each low-resource scenario.

\section{Attention Distributions}
\label{appendix:attn}

\begin{figure*}[!htb]
\centering
\includegraphics[width=1.99\columnwidth]{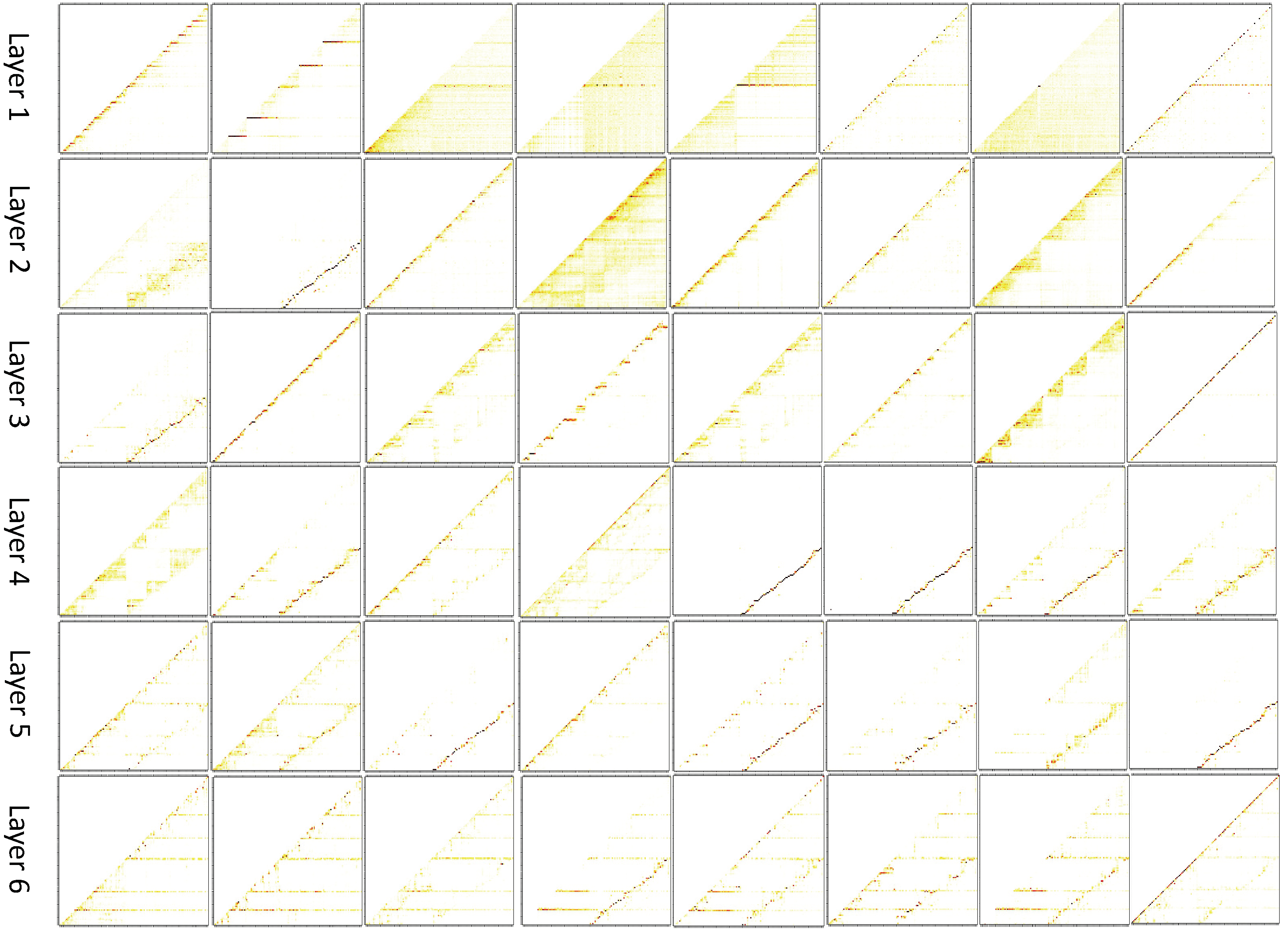} 
\caption{Visualization of all the 48 self attention heads. The x-axis and y-axis are both concatenated source-language summary $S^A$ and target-language summary $S^B$ tokens. Each row contains all of the attention heads of corresponding layer from bottom to the top. The darker color shows the more highly related associations between tokens. 
}
\label{self_all}
\end{figure*}

\begin{figure*}[!htb]
\centering
\includegraphics[width=1.99\columnwidth]{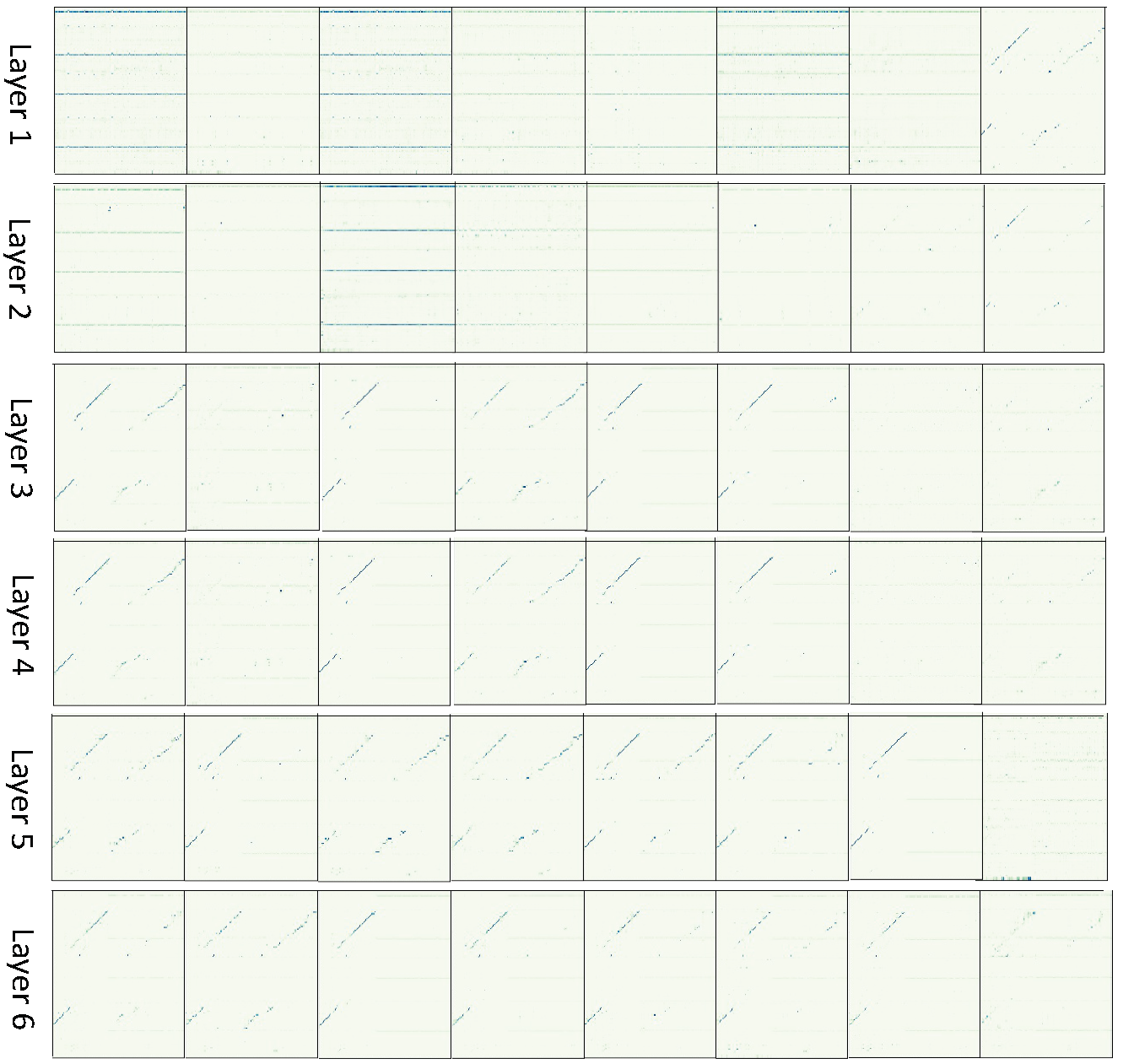} 
\caption{Visualization of all the 48 encoder-decoder attention heads. The x-axis is concatenated source-language summary $S^A$ and target-language summary $S^B$ tokens while the y-axis is document $D^A$ tokens. Each row contains all of the attention heads of corresponding layer from bottom to the top. The darker color shows the more highly related associations between tokens.
}
\label{context_all}
\end{figure*}

In Section ``Probing into Attention Heads'', we selected some representative attention heads. We list all of our trained attention heads among 6 Transformer decoder layers in Figure \ref{self_all} and Figure \ref{context_all} for reference. 

\end{document}